\documentclass[letterpaper]{article} % DO NOT CHANGE THIS
\usepackage{aaai24}  % DO NOT CHANGE THIS
\usepackage{times}  % DO NOT CHANGE THIS
\usepackage{helvet}  % DO NOT CHANGE THIS
\usepackage{courier}  % DO NOT CHANGE THIS
\usepackage[hyphens]{url}  % DO NOT CHANGE THIS
\usepackage{graphicx} % DO NOT CHANGE THIS
\usepackage{natbib}  % DO NOT CHANGE THIS AND DO NOT ADD ANY OPTIONS TO IT
\usepackage{caption} % DO NOT CHANGE THIS AND DO NOT ADD ANY OPTIONS TO IT
\usepackage{algorithm}
\usepackage{algorithmic}

\usepackage{amsmath}
\usepackage{amsfonts} 
\DeclareMathOperator*{\argmax}{arg\,max}
\DeclareMathOperator*{\argmin}{arg\,min}

\usepackage{amsthm}

\usepackage{xcolor}
\usepackage{colortbl}

\usepackage{newfloat}
\usepackage{listings}
\DeclareCaptionStyle{ruled}{labelfont=normalfont,labelsep=colon,strut=off} % DO NOT CHANGE THIS
\floatstyle{ruled}
\newfloat{listing}{tb}{lst}{}
\floatname{listing}{Listing}
\usepackage{bm}
\usepackage{array}

\usepackage{booktabs}
\usepackage{multirow}
\usepackage{subcaption}
\usepackage{xcolor}
\usepackage{colortbl}
\usepackage{pifont}

\title{Sparsity-Guided Holistic Explanation for LLMs with\\ Interpretable Inference-Time Intervention}
\author{
   Zhen Tan\textsuperscript{\rm 1}, Tianlong Chen\textsuperscript{\rm 2}, Zhenyu Zhang\textsuperscript{\rm 3}, Huan Liu\textsuperscript{\rm 1}
}
\affiliations{
    \textsuperscript{\rm 1}Arizona State University\\
    \textsuperscript{\rm 2}University of North Carolina at Chapel Hill\\
    \textsuperscript{\rm 3}University of Texas at Austin
    \\\{ztan36,huanliu\}@asu.edu, tianlong@cs.unc.edu, zhenyu.zhang@utexas.edu
}

\begin{document}

\maketitle

\begin{abstract}
Large Language Models (LLMs) have achieved unprecedented breakthroughs in various natural language processing domains. However, the enigmatic ``black-box'' nature of LLMs remains a significant challenge for interpretability, hampering transparent and accountable applications. While past approaches, such as attention visualization, pivotal subnetwork extraction, and concept-based analyses, offer some insight, they often focus on either local or global explanations within a single dimension, occasionally falling short in providing comprehensive clarity. In response, we propose a novel methodology anchored in sparsity-guided techniques, aiming to provide a holistic interpretation of LLMs. Our framework, termed \textit{SparseCBM}, innovatively integrates sparsity to elucidate three intertwined layers of interpretation: input, subnetwork, and concept levels. In addition, the newly introduced dimension of interpretable inference-time intervention facilitates dynamic adjustments to the model during deployment. Through rigorous empirical evaluations on real-world datasets, we demonstrate that SparseCBM delivers a profound understanding of LLM behaviors, setting it apart in both interpreting and ameliorating model inaccuracies. Codes are provided in supplements.
\end{abstract}

\section{Introduction}

% The arena of natural language processing (NLP) has been profoundly transformed by the rapid progression and sophistication of language models. Pioneered by BERT~\citep{devlin2018bert}, which introduced the transformative power of bidirectional transformers, a new epoch of language modeling was ushered in. Subsequent models, such as RoBERTa~\citep{liu2019roberta}, T5~\citep{raffel2020exploring}, GPT-3~\citep{brown2020language}, GPT-4~\cite{openai2023gpt4}, and specialized models such as BLOOM~\citep{scao2022bloom}, have consistently elevated the benchmark, exhibiting remarkable proficiency in a variety of NLP tasks, such as machine translation and sentiment analysis. 
The advent of Large Language Models (LLMs) has captured the intricacies of language patterns with striking finesse, rivaling, and at times, surpassing human performance~\citep{zhou2022large,openai2023gpt4}. However, their laudable success story is shadowed by a pressing concern: a distinct lack of \textit{transparency} and \textit{interpretability}. As LLMs burgeon in complexity and scale, the elucidation of their internal mechanisms and decision-making processes has become a daunting challenge. The opaque ``black-box'' characteristics of these models obfuscate the transformation process from input data to generated output, presenting a formidable barrier to trust, debugging, and optimal utilization of these potent computational tools. Consequently, advancing the interpretability of LLMs has emerged as a crucial frontier in machine learning and natural language processing research, aiming to reconcile the dichotomy between superior model performance and comprehensive usability. 

\begin{figure}[t]
% \vspace{-0.1cm}
		\centering 
\scalebox{0.58}{
\includegraphics[width=0.8\textwidth]{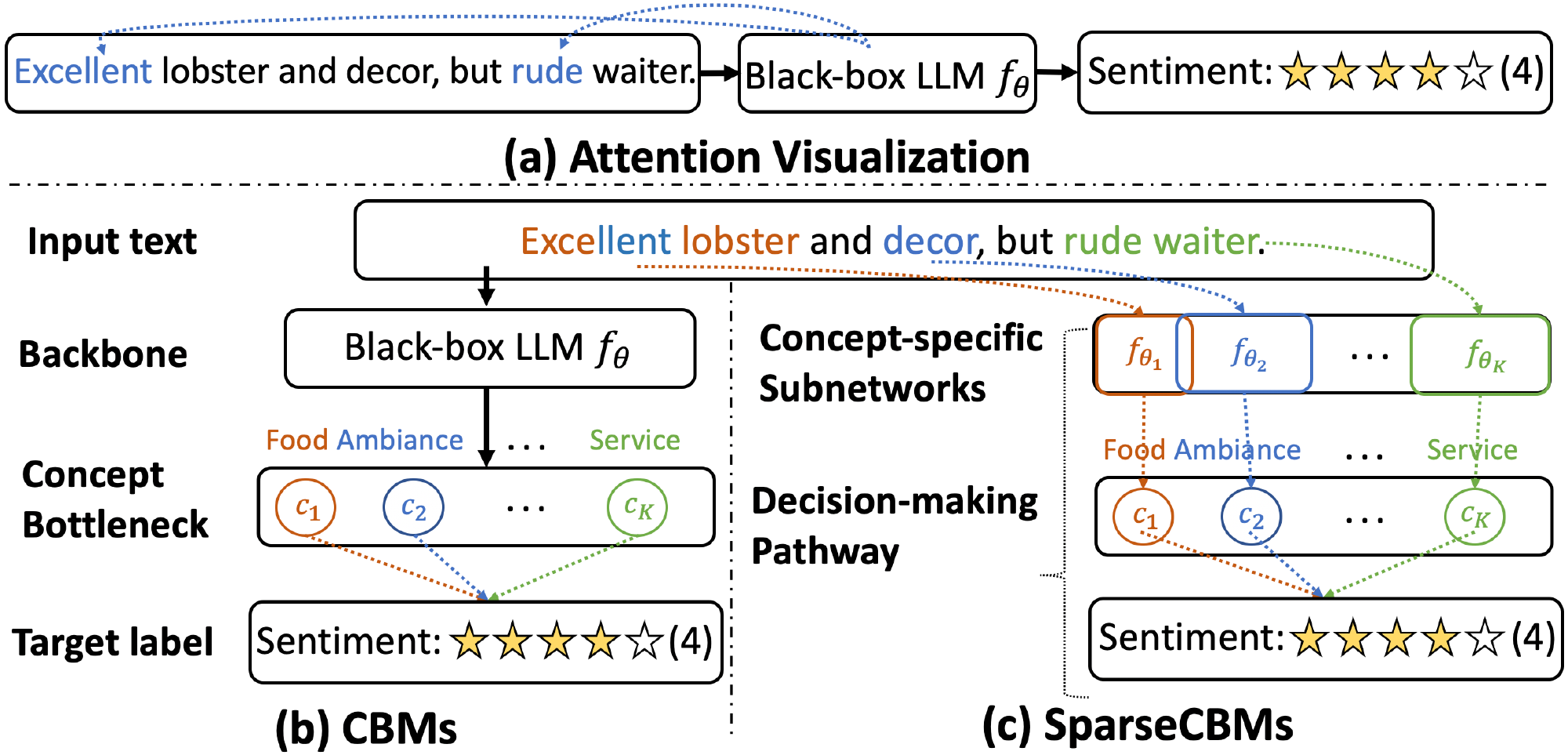}}
	%\subcaptionbox{\includegraphics[width=0.48\textwidth]{img/nwaykshot.pdf}}
% \vspace{-0.3cm}
\caption{The illustration includes:
(a) \textit{Attention visualization} provides a localized, attention-driven explanation. While insightful, this might be less decipherable or intuitive for users outside the realm of computer science.
(b) \textit{CBMs} deliver a broader, concept-level understanding, resonating naturally with human cognition. However, they sometimes miss out on the nuanced, granular insights of the LLM's workings.
(c) \textit{SparseCBMs} outline a holistic decision pathway for each input, seamlessly progressing from tokens, via pertinent subnetworks and concepts, to the final task label. This approach marries the strengths of both local and global explanations, addressing their respective shortcomings.}
% \vspace{-0.3cm}
\label{fig:teaser}
	\end{figure}

The spectrum of interpretability solutions for language models can be broadly bifurcated into two categories. 
\ding{182}~Initial approaches predominantly leverage \textit{local explanations}, employing techniques such as visualization of attention weights~\cite{galassi2020attention}, probing of feature representations~\cite{mishra2017local,lundberg2017unified}, and utilization of counterfactuals~\cite{wu2021polyjuice,ross2021explaining}, among others.
These methods focus on providing explanations at granular levels, such as individual tokens, instances, neurons, or subnetworks, as exemplified in Figure~\ref{fig:teaser}~(a). While these low-level explanations offer a degree of reliability, they often sacrifice \textbf{readability} and \textbf{intuitiveness}~\citep{losch2019interpretability}, thereby constraining their practical applicability. \ding{183}~More recently, researchers have tended to \textit{global explanations}, such as concept-based analyses that inherently resonate with human cognition~\citep{wang2023intepreting,abraham2022cebab}. For instance, one recent work~\citep{tan2023cbm} incorporates Concept Bottleneck Models (CBMs)~\citep{koh2020concept} into pretrained language models, leading to an impressive ``interpretability-utility'' Pareto front. Figure~\ref{fig:teaser} (b) exemplifies this for sentiment analysis tasks, where human-intelligible concepts like ``Food'', ``Ambiance'', and ``Service'' correspond to neurons in the concept bottleneck layer. The final decision layer is designed as a linear function of these concepts, rendering the decision rules easily understandable. However, these methods excessively focus on global explanations. The \textbf{underlying reasoning} between raw input and concepts remains unclear.
% , especially for LLMs hosting billions of parameters.

% . Moreover, interpretability remains underexplored for the realm of large language models, specifically those encompassing billions of parameters, presenting a clear gap in our understanding of these powerful models.

To address these limitations, our work champions a \textit{holistic} interpretation of LLM predictions. We unveil \textit{SparseCBM}, an evolved CBM variant that melds the complementary ``strengths'' of local and global explanations, thereby addressing the individual ``weaknesses'' of each. This confluence is born from rigorous sparsity-guided refinement designed specifically for LLMs, as depicted in Figure~\ref{fig:teaser} (c). Concretely, SparseCBM iteratively prunes the LLM backbone guided by a joint objective of optimizing for both concept and task labels until the desired sparsity level is accomplished. This exercise distills the LLM into distinct yet interconnected subnetworks, each corresponding to a predefined concept. As such, SparseCBM provides a comprehensive and intelligible decision-making pathway for each input text, tracing from tokens through subnetworks and concepts, ultimately leading to the final task label.

Another unique feature is that, SparseCBMs allow \textbf{interpretable inference-time intervention}~\citep{koh2020concept,li2023inference}. The inherent sparsity-driven structure of SparseCBM allows it to adjust its internal parameters dynamically, based on the context of the input. In practical terms, this means that, during inference, SparseCBM can identify potential areas of ambiguity or misconception, and proactively modify its internal decision-making routes without a full-scale retraining. This ``on-the-fly'' adaptability not only enhances prediction accuracy but also offers users a window into how the model adjusts its reasoning in real time. By making these modifications both accessible and understandable, SparseCBM bridges the common chasm between interpretability and agility for LLMs. This real-time decision pathway modification,
% grounded in comprehensible concepts,
stands as a beacon for fostering trust and facilitating more nuanced human-model interactions. In summary, SparseCBM carries the following advantages:
\begin{itemize}
    \item \underline{\textit{Empirical Validation}}: Our experiments reveal that SparseCBM enables interpretability at the token, subnetwork, and concept levels, creating a synergy that surpasses the mere aggregation of these elements.
    \item \underline{\textit{Superior Performance}}: SparseCBM demonstrates state-of-the-art performance on conventional benchmarks, both in terms of concept and task label predictions.
    \item \underline{\textit{Metacognitive Inference-Time Intervention}}: Compared to vanilla CBMs, SparseCBM exhibits a unique capability for efficient and interpretable inference-time intervention. By subtly modulating internal sparsity, SparseCBM learns to sidestep known pitfalls. This property bolsters user trust in SparseCBMs and, by extension, LLMs.
    
    % \item Experimentally, we exihbit that SparseCBM facilitates interpretation at the input, subnetwork, and concept levels, leading to a synergy that transcends the mere aggregation of its individual parts;
    % \item We further demonstrate that SparseCBM obtains state-of-the-art results on standard benchmarks in terms of both concept and task label prediction;
    % \item A unique feature of SparseCBM, compared to conventional CBMs, lies in its capacity for efficient and interpretable inference-time intervention~\citep{koh2020concept,li2023inference}. Through the modulation of internal sparsity, SparseCBM can learn to circumvent previously encountered errors without necessitating retraining of the LLM backbone. This feature serves to bolster the trust of envisioning users in SparseCBM and, by extension, LLMs.
\end{itemize}

\section{Related Work}
\subsection{Interpreting Language Models}
Research on the interpretability of language models has been robust, with previous work focusing on visualization of hidden states and attention weights in transformer-based models~\cite{vig2019multiscale,galassi2020attention}. These techniques, while valuable, often provided granular insights that were not easily interpretable at a high level. Feature importance methods like LIME~\cite{ribeiro2016should} and SHAP~\cite{lundberg2017unified} provided valuable insights into how each input feature contributes to the prediction, but still fail to offer a global understanding of the model behavior, and often lack intuitiveness and readability.

The advent of concept-based interpretability has marked a significant development, offering more global, high-level explanations~\cite{koh2020concept,abraham2022cebab,wang2023intepreting}. Concept Bottleneck Models (CBMs)~\citep{koh2020concept,oikarinenlabel} which incorporate a concept layer into the model, have gained traction recently~\cite{tan2023cbm}. CBMs are trained with task labels and concept labels either \textit{independently}, \textit{sequentially}, or \textit{jointly}. This design enables inference-time debugging by calibrating the activations of concepts. Yet, current CBMs are deficient in their ability to offer granular interpretations, and inference-time interventions remain incapable of altering the language model backbone, leading to recurrent errors. On the other hand, the interpretability of LLMs remains a less explored area. Although some progress has been made, such as guiding LLMs to generate explanations for their predictions using finely tuned prompts~\citep{li2022explanations}, the reliability of these explanations remains questionable. In summary, a reliable method facilitating holistic insights into model behavior is still wanting. In response, our work advances this field by introducing SparseCBM, a holistic interpretation framework for LLMs that tackles both local and global interpretations, thus enhancing the usability and trustworthiness of LLMs.

\subsection{Sparsity Mining for Language Models}
Sparsity-driven techniques, often associated with model pruning, form an energetic subset of research primarily in the pursuit of model compression. At their core, these methods focus on the elimination of less influential neurons while retaining the more critical ones, thereby sustaining optimal model performance~\citep{lecun1990optimal,han2015deep,magnitude,hessian,liu2017learning,he2017channel,zhou2016less}. Contemporary research has shed light on the heightened robustness of pruned models against adversarial conditions, such as overfitting and distribution shifts. 
Typical pruning methods for language models encompass structured pruning~\citep{michel2019sixteen}, fine-grained structured pruning~\citep{lagunas2021block}, and unstructured pruning~\citep{gale2019state}. In brief, unstructured pruning removes individual weights in a network, leading to a sparse matrix, structured pruning eliminates entire structures like neurons or layers for a dense model, while fine-grained structured pruning prunes smaller structures like channels or weight vectors, offering a balance between the previous two. We direct the readers to the benchmark~\cite{liu2023sparsity} for a comprehensive overview. In our case, we focus on unstructured pruning for its effectiveness and better interpretability. 

Recently, studies have underscored the interpretability afforded by sparse networks~\citep{subramanian2018spine}. For instance, \citet{meister2021sparse} delve into the interpretability of sparse attention mechanisms in language models, \citet{liu2022improve} incorporate sparse contrastive learning in an ancillary sparse coding layer to facilitate word-level interpretability, and \citet{oikarinenlabel} demonstrate that a sparsity constraint on the final linear predictor enhances concept-level interpretation of CBMs. Despite their effectiveness, these frameworks restrict sparsity to a handful of layers, leading to unidimensional interpretability that falls short of the desired comprehensiveness. In contrast, our proposed framework, SparseCBM, imposes sparsity across the entire LLM backbone, enabling holistic interpretation at the token, subnetwork, and concept levels.   

\section{Methodology}
\subsection{Preliminary: Concept Bottleneck Models for Language Models
}
\subsubsection{Problem Setup.}\label{sec:setup}
In this study, we aim to interpret the predictions of fine-tuned Large Language Models (LLMs) in text classification tasks. Given a dataset $\mathcal{D} = \{(\bm{x}^{(i)}, y^{(i)}, \bm{c}^{(i)})_{i=1}^N\}$, we consider an LLM $f_{\bm{\theta}}$ that encodes an input text $\bm{x} \in \mathbb{R}^D$ into a latent representation $\bm{z} \in \mathbb{R}^E$, and a linear classifier $g_{\bm{\phi}}$ that maps $\bm{z}$ into the task label $y$.

\subsubsection{Incorporate Concept Bottlenecks for Large Language Models.} Our architecture mainly follows \citet{tan2023cbm}. Instead of modifying LLM encoders, which could significantly affect the quality of the learned text representation, we introduce a linear layer with sigmoid activation $p_{\bm{\psi}}$. This layer projects the learned latent representation $\bm{z} \in \mathbb{R}^E$ into the concept space $\bm{c} \in \mathbb{R}^K$, resulting in a pathway represented as $\bm{x} \rightarrow \bm{z} \rightarrow \bm{c} \rightarrow y$. Here, we allow multi-class concepts for more flexible interpretation. For convenience, we represent CBM-incorporated LLMs as LLM-CBMs (e.g., BERT-CBM). LLM-CBMs are trained with two objectives: (1) align concept prediction $\hat{\bm{c}}=p_{\bm\psi}(f_{\bm\theta}(\bm{x}))$ to $\bm{x}$’s ground-truth concept labels $\bm{c}$, and (2) align label prediction $\hat{y}=g_{\bm\phi}(p_{\bm\psi}(f_{\bm\theta}(\bm{x})))$ to ground-truth task labels $y$. We mainly experiment with our framework optimized through the \textit{joint training} strategy for its significantly better performance, as also demonstrated in~\citet{tan2023cbm}.
Jointly training LLM with the concept and task labels entails learning the concept encoder and label predictor via a weighted sum, $\mathcal{L}_{joint}$, of the two objectives:
\begin{equation} \label{eq:joint}
\begin{aligned}
    \bm{\theta}^{\ast}, \bm{\psi}^{\ast}, \bm{\phi}^{\ast} &= \argmin_{\bm{\theta}, \bm{\psi}, \bm{\phi}} \mathcal{L}_{joint}(\bm{x}, \bm{c}, y) \\ &= \argmin_{\bm{\theta}, \bm{\psi}, \bm{\phi}} [\mathcal{L}_{CE} (g_{\bm{\phi}}(p_{\bm{\psi}}(f_{\bm{\theta}}(\bm{x}),y) \\ &\quad+ \gamma \mathcal{L}_{CE} (p_{\bm{\psi}}(f_{\bm{\theta}}(\bm{x})),\bm{c})].
\end{aligned}
\end{equation} 
It's worth noting that the LLM-CBMs trained jointly are sensitive to the loss weight $\gamma$. We set the default value for $\gamma$ as $5.0$ for its better performance~\citep{tan2023cbm}. Despite the promising progress having been made, present LLM-CBMs typically train all concepts concurrently, leading to intertwined parameters for concept prediction, making the process less transparent and hampering targeted intervention.

\subsection{\textit{SparseCBMs}}
To address the aforementioned issue, the goal of this paper is to provide a holistic and intelligible decision-making pathway for each input text, tracing from tokens through subnetworks and concepts, ultimately leading to the final task label. 
To this end, we introduce \textit{SparseCBM}, a pioneering framework capable of unraveling the intricate LLM architectures into a number of concept-specific subnetworks. Our approach not only outperforms conventional CBMs in concept and task label prediction performance but also proffers enhanced interpretation concerning neuron activations, for instance, illuminating which weights inside the LLM backbone play pivotal roles in learning specific concepts.

Our framework starts with decomposing the joint optimization defined in Eq.~\eqref{eq:joint} according to each concept $c_k$, which is formulated as follows:
\begin{equation}\label{eq:decompose}
\begin{aligned}
    \bm{\theta}^{\ast}, \bm{\psi}^{\ast}, \bm{\phi}^{\ast} &= \{(\bm\theta^{\ast}_k)_{k=1}^{K}\}, \{(\bm\psi^{\ast}_k)_{k=1}^{K}\}, \{(\bm\phi^{\ast}_k)_{k=1}^{K}\} \\ &= \argmin_{\bm\theta, \bm\psi, \bm\phi} \sum_{k=1}^K \mathcal{L}_{joint}(\bm{x}, c_k, y). \\
    &= \argmin_{\bm\theta, \bm\psi, \bm\phi} \sum_{k=1}^K [\mathcal{L}_{CE} (g_{\bm{\phi}_k}(p_{\bm{\psi}_k}(f_{\bm{\theta}}(x),y) \\ &\quad+ \gamma \mathcal{L}_{CE} (p_{\bm{\psi}_k}(f_{\bm{\theta}}(\bm{x})),c_k)],
\end{aligned}
\end{equation}
where $\bm{\phi}_k, \bm{\psi}_k$ are the weights of the $k$th parameter of the projector and classifier, and $\bm{\theta}_k$ is the subnetwork specific for the concept $c_k$, which is explained later. Since both of them are comprised of a single linear layer (with or without the activation function), the involved parameters for $c_k$ can be directly indexed from these models and are self-interpretable~\cite{koh2020concept,tan2023cbm}. 

The remaining task is to excavate concept-specific subnetworks for each concept from the vast architecture of Large Language Models (LLMs). The guiding intuition behind this strategy is to perceive the prediction of concept labels as individual classification tasks, ones that should not strain the entirety of pretrained LLMs given their colossal reserves of knowledge encapsulated in multi-million to multi-billion parameters. We propose an unstructured pruning of the LLM backbone for each concept classification task, such that distinct pruned subnetworks are accountable for different concepts while preserving prediction performance.
% Our experiments show that, a pruned SparseCBM model with $\frac{1}{K}$ parameters can even ourperform its dense counterpart.

\subsubsection{Holistic and Intelligible Decision-making Pathways.}
% This work principally aims to mine concept-specific subnetworks for each concept from the vast architecture of Large Language Models (LLMs). The guiding intuition behind this approach is the interpretation of each concept label prediction as an individual classification task, which should be an endeavor that should pose minimal challenge to pretrained LLMs, given their multimillion to multibillion parameter range. 

We leverage unstructured pruning strategies to carve out concept-specific subnetworks within the LLM backbones. The noteworthy edge of such unstructured pruning strategies lies in their ability to engender weight masks in accordance with the weight importance. Such masks naturally can offer an immediate and clear interpretation. Concretely, we introduce a 0/1 weight mask $M_k$ for each corresponding subnetwork. Consequently, the weights of each subnetwork can be represented as $\bm{\theta}_{\bm{M}_k} = \bm{M}_k \odot \bm{\theta}^\ast$, representing the Hadamard (element-wise) product between the LLM weights $\bm{\theta}^{\ast} \in \mathbb{R}^L$ and the weight mask $\bm{M}_k \in \{0,1\}^L$ for the concept $c_k$.

With well-optimized $\{(\bm{M})_{k=1}^K\}$, during inference, the decision-making pathway can be represented as:
\begin{equation}\label{eq:path}
    \hat{y} = \sum_{k=1}^{K} \bm{\phi}_k^\ast \cdot \sigma(\bm{\psi}_k^\ast \cdot f_{\bm{\theta}_{\bm{M}_k}} (\bm{x})) = \sum_{k=1}^{K} \bm{\phi}_k^\ast \cdot \sigma(\bm{\psi}_k^\ast \cdot f_{\bm{M}_k^ \odot \bm{\theta}^\ast} (\bm{x})),
\end{equation}
where $\sigma(\cdot)$ is the sigmoid activation function of the projector. This decision-making pathway defined in Eq.~\eqref{eq:path} factorizes the parameters of the SparseCBM, and can be optimized through one backward pass of the discomposed joint loss defined in Eq.~\eqref{eq:decompose} with $\bm{\theta}_k^\ast = \bm{\theta}_{\bm{M}_k}$. Importantly, we posit that such decision-making pathways can deliver holistic explanations for the model's predictions. For instance, by scrutinizing the weights in the classifier $g_{\bm{\phi}}$ and the concept activation post the $\sigma$ function, we can get a concept-level explanation regarding the importance of different concepts. Also, visualizing each subnetwork mask $\bm{M}_k$ will furnish a subnetwork-level comprehension of neuron behavior and its importance in acquiring a specific concept and forming predictions. Additionally, the study of the gradient of input tokens in masked concept-specific subnetworks can provide more accurate token-concept mapping. Notably, our experiments demonstrate that SparseCBMs, in addition to providing multi-dimensional interpretations, can match or even surpass their dense counterparts in performance on both concept and task label prediction.
Another unique feature of SparseCBMs lies in that, the weight masks $\{(\bm{M}_k)_{k=1}^K\}$ engendered by unstructured pruning facilitates the process of efficient and interpretable \textit{Sparsity-based Inference-time Intervention}, which is expounded later.

\subsubsection{Concept-Induced Sparsity Mining.}
Next, we elaborate on how to compute those sparsity masks, given an optimized LLM backbone. A second-order unstructured pruning~\cite{hassibi1992second,kurtic2022optimal} for LLMs has been incorporated. Initially, the joint loss $\mathcal{L}$ (we omit the subscript $joint$ for brevity in subsequent equations) can be expanded at the weights of subnetwork $\bm{\theta}_{\bm{M}_k}$ via Taylor expansion:
% \begin{align}
%     \mathcal{L}(\bm{\theta}_{\bm{M}_k}) &\simeq \mathcal{L}(\bm{\theta}^\ast) + (\bm{\theta}_{\bm{M}_k} - \bm{\theta}^\ast)^{\top} \nabla \mathcal{L}(\bm{\theta}^\ast) \nonumber\\&+ \frac{1}{2}(\bm{\theta}_{\bm{M}_k}-\bm{\theta}^\ast)^\top \bm{H}_{\mathcal{L}} (\bm{\theta^\ast})(\bm{\theta_{\bm{M}_k}})
% \end{align}
\begin{equation}
\begin{aligned}
    \mathcal{L}(\bm{\theta}_{\bm{M}_k}) &\simeq \mathcal{L}(\bm{\theta}^\ast) + (\bm{\theta}_{\bm{M}_k} - \bm{\theta}^\ast)^{\top} \nabla \mathcal{L}(\bm{\theta}^\ast) \\& \quad+ \frac{1}{2}(\bm{\theta}_{\bm{M}_k}-\bm{\theta}^\ast)^\top \bm{H}_{\mathcal{L}} (\bm{\theta}^\ast)(\bm{\theta_{\bm{M}_k}} - \bm{\theta}^\ast),
\end{aligned}
\end{equation}
where $\bm{H}_{\mathcal{L}}(\bm{\theta}^\ast)$ stands for the Hessian matrix of the decomposed joint loss at $\bm{\theta}^\ast$. Since $\bm{\theta}^\ast$ is well-optimized, we assume $\nabla \mathcal{L} (\bm{\theta}^\ast) \approx 0$ as the common practice~\cite{hassibi1992second,kurtic2022optimal}. Then, the change in loss after pruning can be represented as:
\begin{equation}
    \Delta \mathcal{L} (\Delta\bm\theta) = \mathcal{L}(\bm{\theta}_{\bm{M}_k}) - \mathcal{L} (\bm\theta^\ast) \simeq \frac{1}{2}\Delta\bm\theta^\top \bm{H}_\mathcal{L} \Delta \bm{\theta},
\end{equation}
where, $\Delta\bm{\theta} = \bm\theta_{\bm{M}_k}-\bm{\theta}^\ast$ signifies the change in LLM weights, that is, pruned parameters. Given a target sparsity $s\in [0,1)$, we seek the minimum loss change incurred by pruning. In our case, the default sparsity is designed as: $s \geq 1-\frac{1}{K}$, implying each subnetwork contains a maximum of $\frac{1}{K}$ parameters in the dense counterpart. Ideally, we desire separate parameters in the LLM backbone to ensure optimal interpretability. Then, the problem of computing the sparsity masks can be formulated as a constrained optimization task:
\begin{equation}\label{eq:optimization}
    \begin{aligned}
        &\min_{\Delta \bm\theta} \quad \frac{1}{2} \Delta \bm{\theta}^\top \bm{H}_{\mathcal{L}} (\bm{\theta^\ast}) \Delta \bm\theta,
        \\& s.t. \quad \bm{e}_b^{\top} \Delta \bm\theta + \theta_b = 0, \quad \forall b \in \bm{Q},
    \end{aligned}
\end{equation}
where $\bm{e}_b$ denotes the $b$th canonical basis vector of the block of weights $\bm{Q}$ to be pruned. 
This optimization can be solved by approximating the
Hessian at $\bm\theta^\ast$ via the dampened empirical Fisher
information matrix~\citep{hassibi1992second,kurtic2022optimal}. Hence, we can derive the optimized concept-specific masks $\{(\bm{M}_k)_{k=1}^K\}$. More details are in Appendix C.

% For convenience, we use BERT models to represent BERT-family models. As described in the previous section, our goal is to extract a dedicated subnetwork from the BERT model for each concept. To achieve this, we first decompose the training with joint loss according to each concept $c_k$ as follows:
% \begin{equation}
%     \theta^{\ast}, \psi^{\ast}, \phi^{\ast} &= \theta^{\ast}, \psi^{\ast}, \phi^{\ast}
    
%     \argmin_{\theta, \psi_k, \phi_k} \mathcal{L}_{joint}(x, c_k, y).
% \end{equation}

% , firstly, we train the BERT model for a specifc concept $c_k, \forall k \in \{1,2,\ldots,K\}$ through the standard joint training strategy. Note that during this training, inside the projector $p_\psi$ and the classifier $g_\phi$, since both of them are linear layers, only the weights corresponding to $c_k$, i.e., $\psi_k, \phi_k$, are updated.
% % The final loss is the sum of losses from all the concepts.
% Then, through the predefined loss $\mathcal{L}_{joint}$, we can get the well-optimized dense BERT model with weights $\theta^{\ast}$:
% \begin{equation}
%     \theta^{\ast}, \psi^{\ast}_k, \phi^{\ast}_k &= \argmin_{\theta, \psi_k, \phi_k} \mathcal{L}_{joint}(x, c_k, y).
% \end{equation}

\subsubsection{Sparsity-based Inference-time intervention.} SparseCBMs also exhibit the capability to allow inference-time concept intervention (a trait inherited from CBMs), thus enabling more comprehensive and user-friendly interactions.  
SparseCBMs allow modulation of the inferred concept activations: $\hat{\bm{a}} = \sigma (p_{\bm\phi}(f_{\bm{\theta}}(\bm{x})))$. There are two straightforward strategies for undertaking such intervention. The first option is the \textit{oracle intervention}~\citep{koh2020concept}, where human experts manually calibrate the concept activations $\hat{\bm{a}}$ and feed them into the classifier. 
% In practical scenarios, domain experts interacting with the model might intervene to ``correct'' any potentially erroneous concept activations predicted by the model.
Despite its apparent simplicity, oracle intervention directly operates on concept activations and, therefore, cannot fix the flawed mapping learned by the LLM backbone. As a consequence, the model will replicate the same error when presented with the same input. Meanwhile, another strategy involves further fine-tuning the LLM backbone on the test data. However, this approach is not only inefficient but also has a high risk of leading to significant overfitting on the test data. Those limitations present a barrier to the practical implementation of CBMs in high-stakes or time-sensitive applications.  

As a remedy, we further propose a \textit{sparsity-based intervention} that is self-interpretable and congruent with SparseCBMs. It helps LLMs to learn from each erroneously predicted concept during inference time, while preserving overall performance. The core idea is to subtly modify the concept-specific masks for the LLM backbone when a mispredicted concept is detected. Specifically, parameters of the LLM backbone $f_\theta$, projector $p_\psi$, and the classifier $g_\phi$ are frozen, while the concept-specific masks $\{(\bm{M}_k)_{k=1}^K\}$ is kept trainable. During the test phase, if a concept prediction $\hat{c}_k$ for an input text $\bm{x}$ is incorrect, we acquire the gradient ${\mathcal{G}_k}(\bm{x})$ for the corresponding subnetwork $f_{\bm{\theta}_{\bm{M}_k}}$, and modulate the learned mask $\bm{M}_k$ accordingly. 

Inspired by~\citet{evci2020rigging,sun2023simple}, we define the saliency scores for LLM parameters by the $l_2$-norm of the product of the gradient of the mask and the parameter weights: $\mathcal{S} = \|\mathcal{G}_k(\bm{x}) \cdot \bm{\theta}^\ast \|$. Subsequently, we perform the following two operations based on the saliency scores: (1)~\textbf{Drop} a proportion of $r$ unpruned weights with the lowest saliency scores: $\argmin_m^{r\cdot |\bm{\theta}|} \mathcal{S}_m$, $\forall m \in |\bm{\theta}_{\bm{M}_k}|$. (2) \textbf{Grow} a proportion of $r$ pruned weights with the highest saliency scores: $\argmax_m^{r\cdot |\bm{\theta}|} \mathcal{S}_m$, $\forall m \in |\bm{\theta} \setminus \bm{\theta}_{\bm{M}_k}|$. Here $m$ refers to the parameter index of the LLM backbone. By dropping and growing an equal number of parameters, the overall sparsity $s$ of the LLM backbone remains unchanged. This mask-level intervention is further optimized through the decomposed joint loss $\mathcal{L}_{joint}$ defined in Eq.~\eqref{eq:decompose}. Note that $r$ is set as a relatively small value (e.g., 0.01) to compel the model to retain the overall performance while learning from the mistake. Our experiments validate that the proposed sparsity-based intervention can effectively enhance inference-time accuracy without necessitating training of the entire LLM backbone. Also, the intervened parameters provide insight into the parameters that contributed to each misprediction.

\section{Experiments}
\subsection{Experimental Setup}
\subsubsection{Datasets.} Our experiments are conducted on two widely-used real-world datasets: \texttt{CEBaB}~\citep{abraham2022cebab} and \texttt{IMDB-C}~\cite{tan2023cbm}. Each of them is a text-classification dataset comprised of human-annotated concept and task labels. Their statistics are presented in Table~\ref{tab:data}.
\subsubsection{LLM backbones.}
In this research, we primarily consider two widely-recognized, open-source lineages of pretrained LLMs: the BERT-family models~\cite{devlin2018bert,liu2019roberta,sanh2019distilbert} and OPT-family models~\cite{zhang2022opt}. Specially, we also include directly prompting GPT4~\cite{openai2023gpt4} as a baseline to let it generate concept and task labels for given texts.
% Used prompts and more details are included in Appendix B. 
Even though being proprietary, GPT4 is widely regarded as the most capable LLM currently, so we choose it as the baseline backbone. For better performance, we obtain the representations of the input texts by pooling the embedding of all tokens. Reported scores are the averages of three independent runs. Our work is based on general text classification implementations. The PyTorch Implementation is available at \textit{https://github.com/Zhen-Tan-dmml/SparseCBM.git}.

\begin{table}[t]
% \scalebox{0.69}{
\centering
\setlength{\tabcolsep}{1pt} % Change 6pt to your desired value
\fontsize{9pt}{10pt}\selectfont
\begin{tabular}{@{}c|cccc@{}}
\toprule
\multirow{2}{*}{\textbf{Dataset}} & \multicolumn{4}{c}{\texttt{CEBaB}    (5-way classification)}                                   \\ \cmidrule(l){2-5} 
                                  & \multicolumn{2}{c|}{\textbf{Train/Dev/Test}} & \multicolumn{2}{c}{1755 / 1673 / 1685} \\ \midrule
\multirow{5}{*}{\textbf{Concept}} & \textbf{Concept}      & \textbf{Negative}    & \textbf{Positive}  & \textbf{Unknown}  \\ \cmidrule(l){2-5} 
                                  & Food                  & 1693 (33\%)        & 2087 (41\%)      & 1333 (26\%)     \\
                                  & Ambiance              & 787 (15\%)         & 994 (20\%)       & 3332 (65\%)     \\
                                  & Service               & 1249 (24\%)        & 1397 (27\%)      & 2467 (49\%)     \\
                                  & Noise                 & 645 (13\%)         & 442 (9\%)        & 4026 (78\%)     \\ \midrule
\multirow{2}{*}{\textbf{Dataset}} & \multicolumn{4}{c}{\texttt{IMDB-C}    (2-way classification)}                                    \\ \cmidrule(l){2-5} 
                                  & \multicolumn{2}{c|}{\textbf{Train/Dev/Test}} & \multicolumn{2}{c}{100 / 50 / 50}      \\ \midrule
\multirow{5}{*}{\textbf{Concept}} & \textbf{Concept}      & \textbf{Negative}    & \textbf{Positive}  & \textbf{Unknown}  \\ \cmidrule(l){2-5} 
                                  & Acting                & 76 (38\%)            & 66 (33\%)          & 58 (29\%)         \\
                                  & Storyline             & 80 (40\%)            & 77 (38\%)        & 43 (22\%)       \\
                                  & Emotion     & 74 (37\%)            & 73 (36\%)        & 53 (27\%)       \\
                                  & Cinematography        & 118 (59\%)           & 43 (22\%)        & 39 (19\%)       \\ \bottomrule
\end{tabular}
% \vspace{-2mm}
\caption{Statistics of experimented datasets and concepts.}\label{tab:data}
\end{table}

\begin{table}[t]
% \vspace{-0.3cm}
\centering
% \vspace{-4mm}
\setlength{\tabcolsep}{1pt} % Change 6pt to your desired value
\fontsize{9pt}{9.5pt}\selectfont
\begin{tabular}{cccccc}
\toprule
\multirow{2}{*}{\textbf{Backbone}}   & \multirow{2}{*}{\textbf{Acc. / F1}} & \multicolumn{2}{c}{\texttt{CEBaB}}                                & \multicolumn{2}{c}{\texttt{IMDB-C}}                                 \\ \cline{3-6} 
                                     &                                     & \textbf{Concept}                & \textbf{Task}                   & \textbf{Concept}                & \textbf{Task}                   \\ \midrule
\textbf{GPT4}                        & Prompt                           & \multicolumn{1}{l}{75.9/71.5} & \multicolumn{1}{l}{51.3/45.9} & \multicolumn{1}{l}{64.5/61.5} & \multicolumn{1}{l}{71.4/68.7} \\ \midrule
\multirow{3}{*}{\textbf{DistilBERT}} & Standard                            & -                               & 70.3/80.4                     & -                               &     77.1/73.8                           \\
                                     & CBM                                 & 81.1/83.5                     &  63.9/76.5                     &        67.5/63.8                         & 76.5/69.8                                \\
                                     & SparseCBM                           & \textbf{82.0/84.0}            & \textbf{64.7/77.1}            &     \textbf{68.4/64.3}                            &    \textbf{76.9/71.4}                             \\ \midrule
\multirow{3}{*}{\textbf{BERT}}       & Standard                            & -                               & 67.9/79.8                     & -                               &  78.3/72.1                               \\
                                     & CBM                                 & 83.2/85.3                     & 66.9/78.1                     &   68.2/62.8                              &    77.3/70.4                             \\
                                     & SparseCBM                           & \textbf{83.5/85.6}            & \textbf{66.9/79.1}            &    \textbf{69.8/65.2}                             &    76.5/\textbf{71.6}                             \\ \midrule
\multirow{3}{*}{\textbf{RoBERTa}}    & Standard                            & -                               & 71.8/81.3                     & -                               &   82.2/77.3                              \\
                                     & CBM                                 & 82.6/84.9                     & 70.1/81.3                     &       69.9/68.9                          &     81.4/79.3                            \\
                                     & SparseCBM                           & \textbf{82.8/85.5}            & \textbf{70.3/81.4}            &    \textbf{70.2/69.7}                             &    \textbf{81.5/79.9}                             \\ \midrule
\multirow{3}{*}{\textbf{OPT-125M}}   & Standard                            & -                               & 70.8/81.4                     & -                               &    84.3/80.0                             \\
                                     & CBM                                 & 85.4/87.3                     & 68.9/79.7                     &   68.7/66.5                              &      81.8/78.2                           \\
                                     & SparseCBM                           & \textbf{86.2/88.0}            & \textbf{68.9/79.8}            &    \textbf{70.0/67.4}                             &     \textbf{82.6/79.9}                            \\ \midrule
\multirow{3}{*}{\textbf{OPT-350M}}   & Standard                            & -                               & 71.6/82.6                     & -                               &        86.4/83.5                         \\
                                     & CBM                                 & 87.8/89.4                     & 69.9/80.7                     &   72.6/70.5                              &      84.5/82.4                           \\
                                     & SparseCBM                           & {87.3/88.7}            & {68.2/79.8}            &     \textbf{73.3/71.1}                            &     \textbf{85.0/82.5}                            \\ \midrule
\multirow{3}{*}{\textbf{OPT-1.3B}}   & Standard                            & -                               & 74.7/83.9                     & -                               &       88.4/83.7                          \\
                                     & CBM                                 & 90.0/91.5                     & 73.6/82.1                     &   76.8/74.6                              &      85.7/83.3                           \\
                                     & SparseCBM                           & 89.9/\textbf{91.6}            & \textbf{73.8/82.6}            &   76.4/\textbf{74.7}                              &      \textbf{86.6/83.9}                           \\ \bottomrule
\end{tabular}
% \vspace{-2mm}
\caption{Comparisons of task accuracy and interpretability using \texttt{CEBaB} and \texttt{IMDB-C} datasets with BERT-family and OPT-family models as the backbones. Metrics for both task and concept labels are \textit{Accuracy}/\textit{Macro F1} in $\%$. A score in \textbf{bold} indicate that the SparseCBM under the current setting outperforms its dense CBM counterpart.}\label{tab:compare}
\end{table}

\subsection{Interpretability}
\subsubsection{Utility v.s. Interpretability.}
Table~\ref{tab:compare} presents the performance of the concept and task label prediction:
\begin{itemize}
    \item \textbf{Multidimensional Interpretability:} SparseCBMs stand out by offering multidimensional interpretability without compromising task prediction performance. In comparison with standard LLMs (which are fine-tuned exclusively with task labels), SparseCBMs grant concept-level interpretability with only a slight dip in task prediction accuracy. Impressively, SparseCBMs can match or even outperform their dense CBM counterparts while providing multifaceted explanations that extend beyond mere concepts. This underlines the potency of SparseCBMs in striking a balance between interpretability and utility.
    \item \textbf{Scalability with Larger LLM Backbones:} The utilization of larger LLMs within SparseCBMs leads to superior interpretability-utility Pareto fronts. This observation validates our guiding hypothesis that predicting concept labels should not strain the entirety of pretrained LLMs as they are individual classification tasks. Larger LLMs, being repositories of more knowledge through increased parameters, facilitate easier pruning.
    % We discuss the potential of larger LLMs for pruning in sensitivity analyses.
    \item \textbf{Limitations of Direct Prompting:} When directly prompting LLMs, such as GPT4 (without fine-tuning on the target datasets), to predict concept and task labels, the resulting performance is noticeably inferior. This highlights the necessity of learning concepts and task labels in target domains. Additionally, since LLMs' task predictions are autoregressively generated and do not rely entirely on the generated concepts, doubts arise regarding the reliability of concept-level explanations. 
    % This further validates the necessity of this research.
\end{itemize}

\begin{figure}[b]
% \vspace{-0.2cm}
  \centering
  \scalebox{1}{
                  \includegraphics[width=\linewidth]{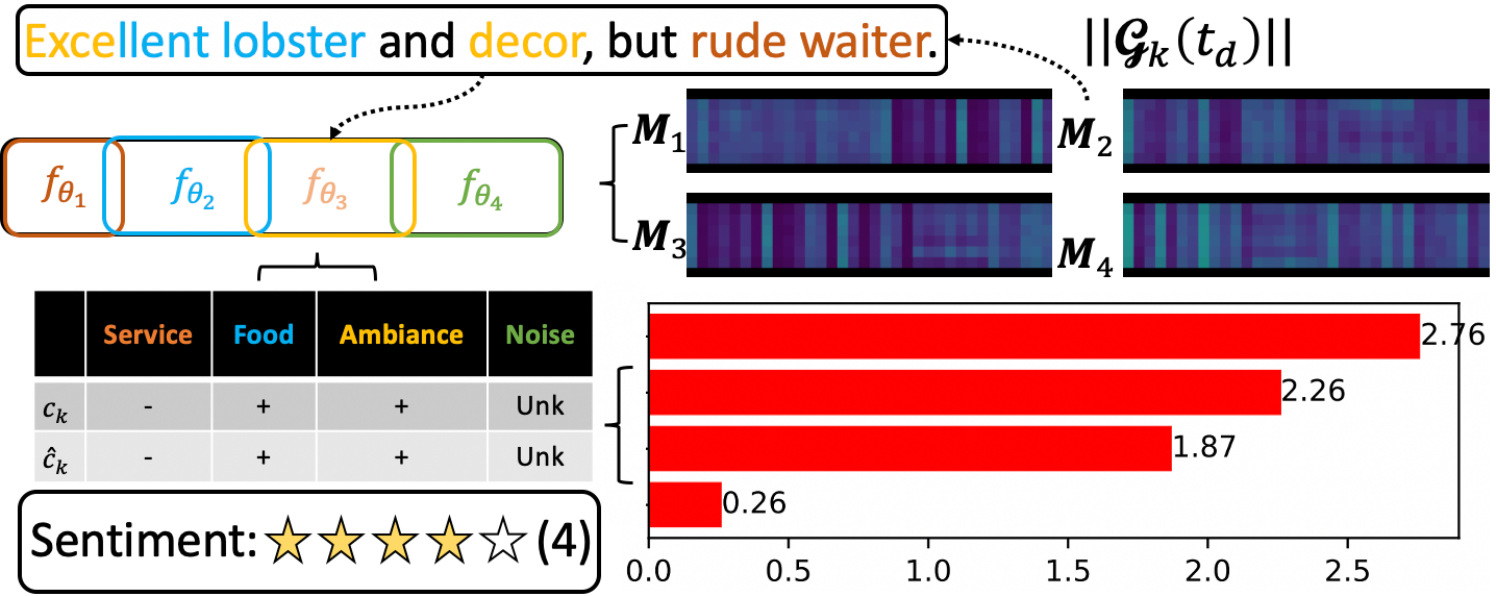}}
  % \vspace{-1mm}
  \caption{{The illustration of a decision pathway of a toy example from the SparseCBM framework with BERT as the backbone. The binary weight masks for each concept is represented as a heatmap.}}\label{fig:example}
  % \vspace{-4mm}
\end{figure}

\subsubsection{Explainable Prediction Pathways.}
The centerpiece of this paper revolves around providing a transparent and interpretive decision-making pathway for each input text vector $\bm{x} = [t_1,t_2,\cdots, t_d,\cdots,t_D]$, where $t_d, \forall d \in D$ denotes the tokens in the input text. SparseCBMs, at inference time, unravel the following layers of understanding across the decision-making trajectory:

\begin{enumerate}
\item \textbf{Subnetwork-Level Explanation:} Identification of specific neurons within the LLM backbones responsible for corresponding concepts. This insight is achieved by visualizing individual binary subnetwork masks
$\bm{M}_k$.
\item \textbf{Token-Level Explanation:} Detection of the tokens instrumental in shaping a particular concept. This analysis is carried out by evaluating the gradient of each subnetwork mask with respect to individual tokens $\|\mathcal{G}_k(t_d)\|$.
\item \textbf{Concept-Level Explanation:} Understanding the predicted concept labels $\hat{c}_k$ and their contribution to the final prediction. This is captured by computing the dot product between each predicted concept activation and the corresponding weight of the linear predictor: $\bm{\phi}_k\cdot\hat{\bm{a}}_k$.
\end{enumerate}
% (1) Subnetwork-level explanation of which neurons in the LLM backbones are responsible for learning the corresponding concepts. This is achieved by visualizing each binary subnetwork mask $\bm{M}_k$ (2) Token-level explanantion on which tokens are responsible for the specific concept. This is achieved by measuring the gradient of each subnetwork mask towards each token $\|\mathcal{G}_k(t_d)\|$. (3) Concept-level explanantion on predicetd concept labels $\hat{c}_k$ and how concepts contribute to the final prediction. This is achieved by measuring the dot product of each predicted concept activation and the corresponding weight of the linear predictor: $\bm{\phi}_k\cdot\hat{\bm{a}}_k$. 
A schematic representation of the decision-making process for a representative example is provided in Figure~\ref{fig:example}, with ``Neg Concept'' denoting negative concept values. Additional real-world examples are delineated in Appendix~D. Several interesting findings can be drawn from those results:
\begin{itemize}
\item \textbf{Neural Responsibility Across Concepts:} Various concepts necessitate differing proportions of neurons in the LLM backbone for learning. This resonates with our ambition to demystify the ``black-box'' LLM backbones by partitioning them into distinct subnetworks, each accountable for an individual concept. Intriguingly, overlaps exist among subnetworks, reflecting that strict disentanglement constraints were not imposed on the backbone parameters. This opens avenues for future research into entirely concept-wise disentangled LLM backbones.
\item \textbf{Holistic Decision Pathway:} The SparseCBM framework successfully crafts a comprehensive decision-making path that navigates from tokens, through subnetworks and concepts, culminating in the final task label. This rich interpretability paves the way for unique insights into practical applications. For instance, although concepts like ``Food'' and ``Ambiance'' may carry identical positive values, the ``Food'' concept may wield greater influence on the final task label. Additionally, careful examination of parameter masks can shed light on the root causes behind mispredicted concepts, enabling effective and interpretable interventions. We explore this topic further in the subsequent section.
\end{itemize}

\subsection{Inference-time Intervention}
\begin{figure}[b]
% \vspace{-0.15cm}
		\centering
\captionsetup[sub]{skip=-0.5pt}
\subcaptionbox{Concept Prediction}
{\includegraphics[width=0.225\textwidth]{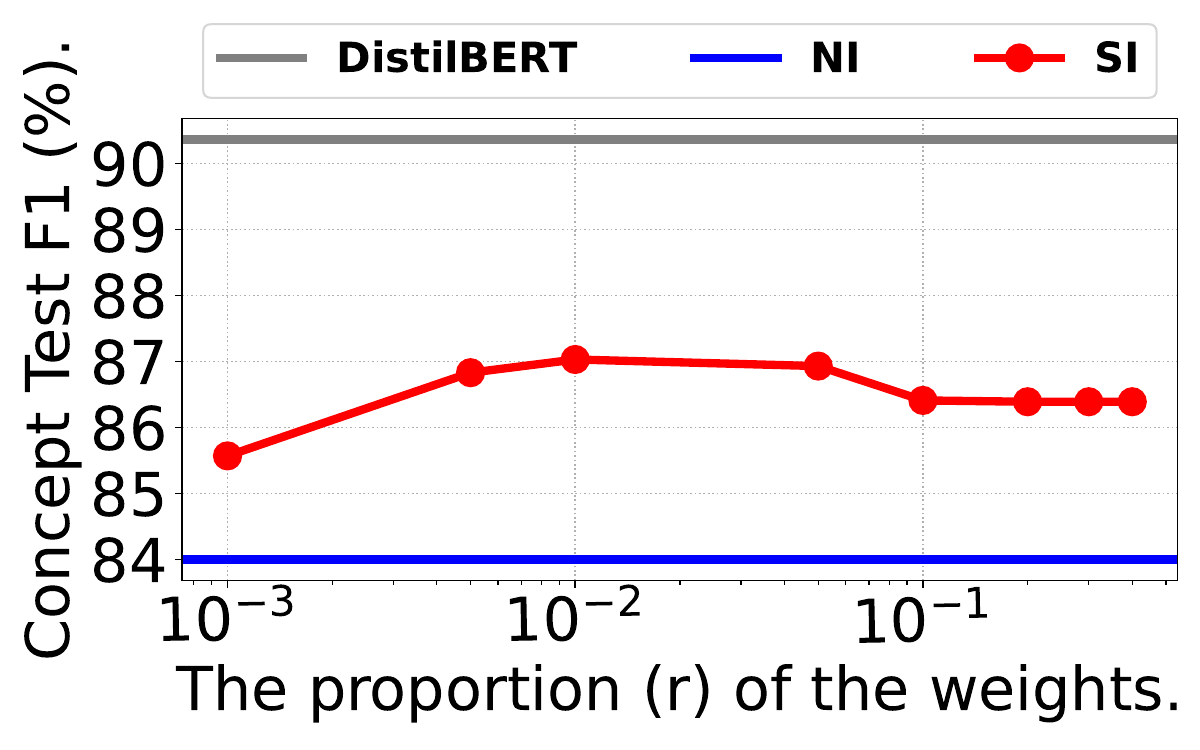}}
\subcaptionbox{{Task Prediction}}
{\includegraphics[width=0.225\textwidth]{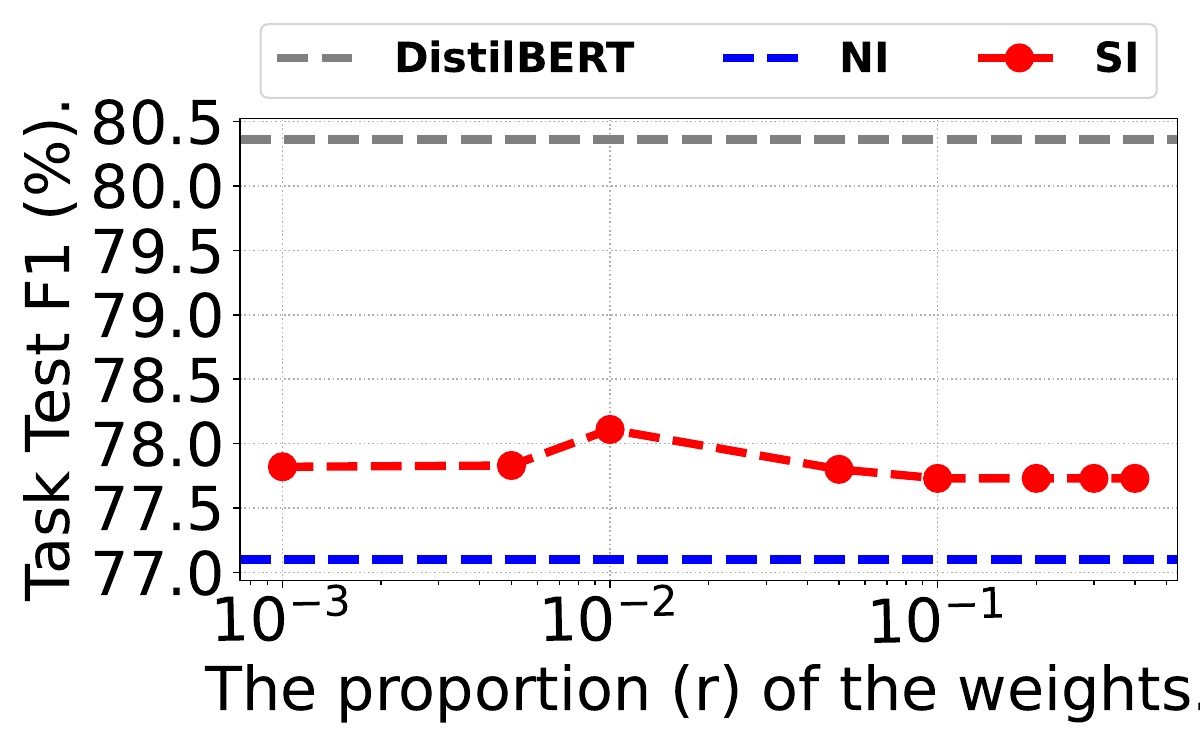}}
% \vspace{-3mm}
		\caption{The results of Test-time Intervention. ``NI'' denotes ``no intervention'', ``SI'' denotes ``Sparsity-based Intervention''. (a) and (b) represent the results for concept and task label prediction respectively. The x-axis indicates the proportion ($r$) of the weights to perform the intervention.}
		\label{fig:intervene}
% \vspace{-0.11in}
	\end{figure}
SparseCBMs distinguish themselves by enabling sparsity-based inference-time intervention, compared to vannila CBMs. This innovative feature creates a pathway for more refined, user-centric interactions by subtly adjusting the masks without the need for direct retraining of the LLM backbone. The significance of this intervention approach lies in its application to real-world scenarios where users often find it easier to articulate broad concepts (e.g., food quality) rather than precise sentiment scores or categorical labels.

\subsubsection{Experimental Evaluation.}
To methodically evaluate this intervention strategy, extensive experiments were conducted on the \texttt{CEBaB} dataset, employing DistilBERT as the representative LLM backbone. The insights gleaned from these experiments apply consistently to other LLMs as well. Figure~\ref{fig:intervene} provides a detailed comparison between concept and task label predictions using SparseCBMs against a baseline, where a vanilla DistilBERT is independently trained to classify concept or task labels. These baseline scores serve as a theoretical upper bound for prediction accuracy, providing a reliable and illustrative benchmark.
This analytical exploration not only validates the proposed sparsity-based intervention's efficacy in enhancing inference-time accuracy for both concept and task predictions but also reveals its elegance in execution. With minimal alterations to the underlying model structure, remarkable improvements are achieved. Even for a relatively small model, DistilBERT, the optimal adjustment proportion is found to be a mere $1\%$, translating to modifications in only $2\%$ of the backbone parameters.

\begin{figure}[t]
% \vspace{-0.2cm}
  \centering
  \scalebox{0.95}{
  % \centering
  \includegraphics[width=\linewidth]{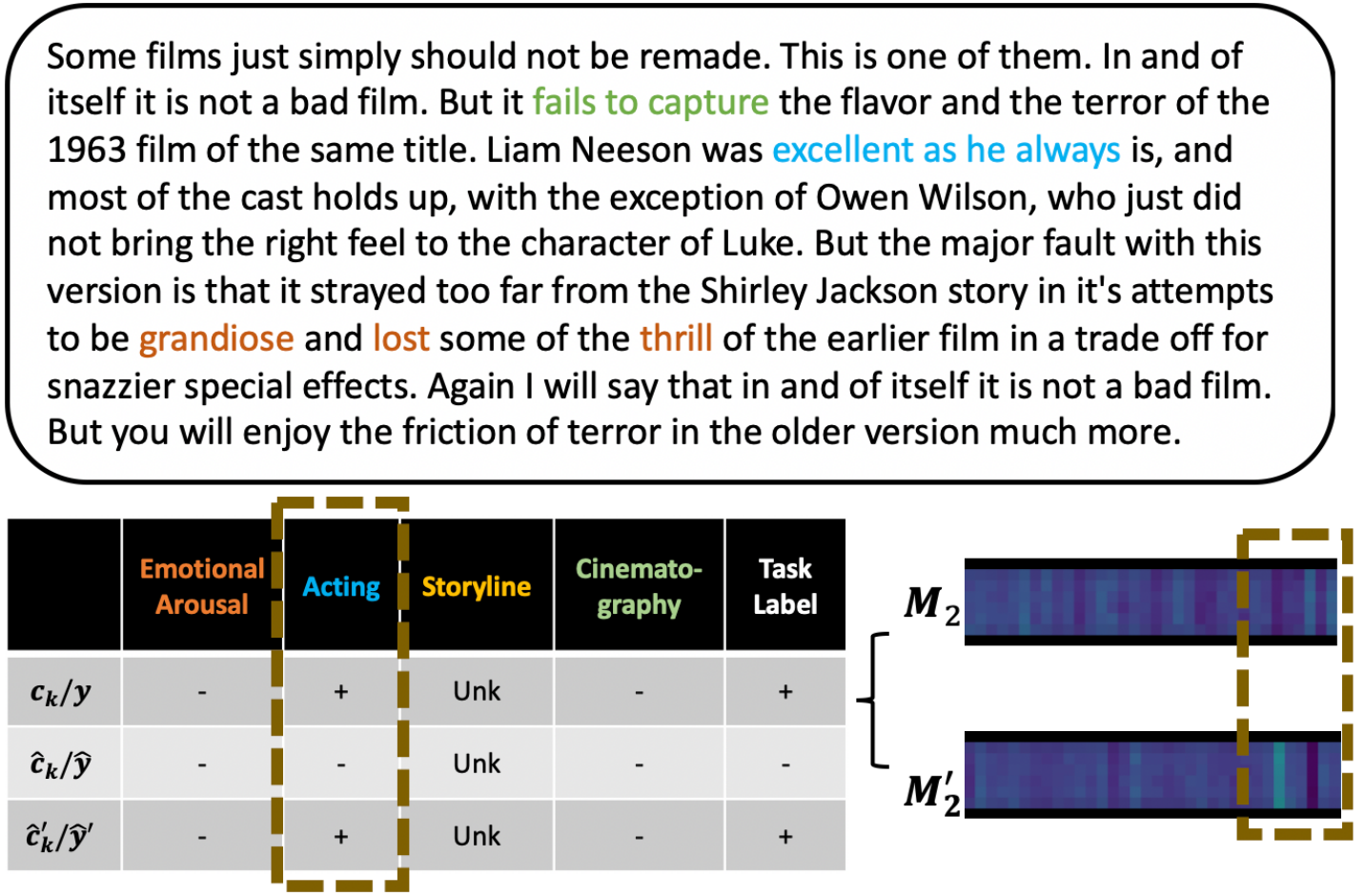}}
  \caption{Illustration of the explainable prediction for a real-world example from the \texttt{IMDB-C} dataset using OPT-350m as the backbone. The brown boxes with dash lines indicate the test-time intervention on corresponding concepts by modulating the corresponding mask. $\bm{M}_2$ and $\bm{M}_2^\prime$ denote the parameter masks for the second concept, ``Acting'', before and after the intervention, respectively. We visualize $\bm{M}_2^\prime$ after seeing all test samples.}
  \label{fig:inter}
  % \vspace{-0.4cm}
\end{figure}

\subsubsection{Robustness and Adaptability.}
These results shed light on the broader applicability and resilience of sparsity-based intervention across various contexts and domains. The capacity to implement such nuanced adjustments without the resource-intensive process of retraining the entire model marks a substantial advancement toward more agile, responsive machine learning systems. This adaptability resonates with the growing demand for models that can quickly adapt to ever-changing requirements without compromising on performance or interpretability.

\subsubsection{Case Study and Insights.}
To provide an in-depth illustration, a case study depicting the sparsity-based intervention process is presented in Figure~\ref{fig:inter}. This visualization elucidates how the predicted label for the concept ``Acting'' can be transformed from incorrect `` -'' to correct ``+'', subsequently refining the final task label. But the insights run deeper: by visualizing the parameter masks before ($\bm{M}_2$) and after ($\bm{M}_2^\prime$) the intervention, we expose the neural mechanics behind the misprediction and the corrective strategy at the neuron level.
This ability to not only correct but also interpret the underlying reasons for prediction errors enhances the overall trustworthiness and usability of the model. In conjunction with the experimental findings, this case study amplifies our understanding of the potential for sparsity-based interventions, not merely as a method for model fine-tuning, but as a principled approach towards more transparent and adaptable AI systems.

\subsubsection{Implication.}
The integration of sparsity-based inference-time intervention within SparseCBMs represents a confluence of accuracy, flexibility, and interpretability. Through careful experimentation and insightful case studies, this work lays the groundwork for models that respond dynamically to the needs of users, augmenting human-machine collaboration in complex decision-making processes. It is a promising step towards building AI models that are not only more effective but also more aligned with the human-centered objectives and ethical considerations of modern machine learning applications.

\subsection{Sensitivity Analysis on the Sparsity $s$}
\begin{figure}[t]
% \vspace{-0.2cm}
  % \centering
  \scalebox{1}{
  \includegraphics[width=\linewidth]{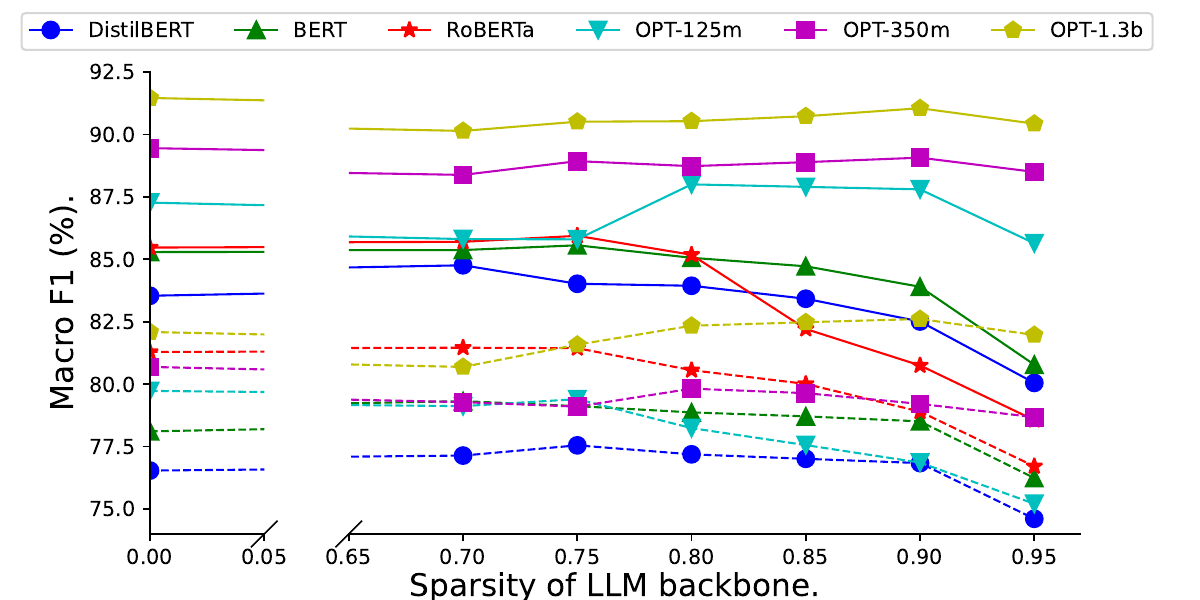}}
  \caption{The performance of SparseCBMs across varying LLM backbones in relation to the target sparsity $s$ on the \texttt{CEBaB} dataset. Solid lines delineate scores for concept label predictions. Dashed lines capture those for task label predictions. Notably, larger LLM backbones are adept at handling increased sparsity without compromising on prediction efficacy. Nonetheless, excessive pruning invariably impinges on the performance across all LLM backbones.}
  \label{fig:logit_0}
  % \vspace{-0.4cm}
\end{figure}

In Figure~\ref{fig:logit_0}, we study the effect of target Large Language Model (LLM) sparsity on concept and task prediction performance across various LLM sizes. The results reveal an interesting trend: larger LLMs tend to have a higher optimal sparsity level compared to smaller ones. This is attributed to the greater knowledge repository and higher redundancy present in larger LLMs, allowing for more extensive pruning without significant performance loss.
However, a delicate balance must be struck. While larger LLMs can accommodate more pruning, overdoing it may harm performance. Identifying this balance remains an intriguing avenue for future research, as well as investigating how different pruning strategies interact with various tasks and data distributions.    

\section{Conclusion}
In this study, we introduced \textit{Sparse Concept Bottleneck Models} (SparseCBMs), a novel method integrating the interpretability of Concept Bottleneck Models (CBMs) with the efficiency of unstructured pruning. By exploiting the properties of second-order pruning, we constructed concept-specific sparse subnetworks in a Large Language Model (LLM) backbone, thereby providing multidimensional interpretability while retaining model performance.
Additionally, we proposed a sparsity-driven inference-time intervention mechanism that improves accuracy at inference time, without the need for expensive fine-tuning LLMs. This intervention mechanism effectively identifies the parameters that contribute to each misprediction, enhancing interpretability further.
Through rigorous experiments, we demonstrated that SparseCBMs match the performance of full LLMs while offering the added benefits of increased interpretability. Our work underscores the potential of sparsity in LLMs, paving the way for further exploration of this intersection. We envisage future investigations to refine the use of structured sparsity, such as group or block sparsity, to further enhance model transparency and efficiency.

\section*{Ethical Statement}
This research explores methods to enhance the interpretability and reliability of large language models (LLMs) through the proposed Sparse Concept Bottleneck Models (SparseCBMs). While the development and application of such technology have benefits, including improved model understanding, and more efficient use of computational resources, several considerations arise that warrant discussion.

\textit{Transparency and Explainability:} Though our work aims to make models more interpretable, the actual understanding of these models can still be quite complex and may be beyond the reach of the general public. Furthermore, the opacity of these models can potentially be exploited, reinforcing the need for ongoing work in model transparency.

\textit{Robustness:} As indicated in~\citep{tan2023cbm,wang2023noise}, the proposed framework is sensitive to the noisy concept and target task labels, requesting future work in model robustness. Potential direction include selective learning~\citep{li2023disc,li2023csgnn}, knowledge editting~\citep{wang2023knowledge}, to name a few.

\textit{Efficiency:} It is worthnoteing that, even though the inference-time intervention is highly efficient, SparseCBM require more training time due to the cocnept-specific pruning. Potential way to enhance the training efficiency is to share part of the sparsity among concepts, as studied in~\citep{wang2020learn,chen2021long}.

\textit{Label Reliance:} SparseCBMs, along with other CBM variants, necessitate the annotation of concepts. To reduce this burden, several approaches are promising. These include leveraging other LLMs for automated annotation, as discussed in~\citep{tan2023cbm,wang2023contrastive}, employing data-efficient learning techniques~\citep{tan2022graph}, and exploring the acquisition of implicit concepts through dictionary learning methods~\citep{wang2022neural}.

\textit{Misuse:} Advanced AI models like LLMs can be repurposed for harmful uses, including disinformation campaigns or privacy infringement~\cite{jiang2023disinformation,chen2023combating}. It's crucial to implement strong ethical guidelines and security measures to prevent misuse.

\textit{Automation and Employment:} The advancements in AI and machine learning could lead to increased automation and potential job displacement. We must consider the societal implications of this technology and work towards strategies to manage potential employment shifts.

\textit{Data Bias:} If the training data contains biases, LLMs may amplify these biases and result in unfair outcomes. We need to continue to develop methods to mitigate these biases in AI systems and promote fair and equitable AI use.

In conducting this research, we adhered to OpenAI's use case policy and are committed to furthering responsible and ethical AI development. As AI technology advances, continuous dialogue on these topics will be needed to manage the potential impacts and ensure the technology is used for the betterment of all.

\section*{Acknowledgments}
This work is supported by the National Science Foundation (NSF) under grants IIS-2229461.

% \bigskip
% \noindent Thank you for reading these instructions carefully. We look forward to receiving your electronic files!

\bibliography{aaai24}
\newpage
\section{A. Definitions of Training Strategies}
\label{app:def}

Given a text input $x \in \mathbb{R}^D$, concepts $c\in \mathbb{R}^K$ and its label $y$, the strategies for fine-tuning the text encoder $f_\theta$, the projector $p_\psi$ and the label predictor $g_\phi$ are defined as follows:

\noindent\textit{i) Vanilla fine-tuning an LLM:} The concept labels are ignored, and then the text encoder $f_\theta$ and the label predictor $g_\phi$ are fine-tuned either as follows:
\begin{equation*}
    \theta, \phi = \argmin_{\theta, \phi} \mathcal{L}_{CE} (g_\phi(f_\theta(x), y),
\end{equation*}
or as follows (frozen text encoder $f_\theta$):
\begin{equation*}
    \phi = \argmin_{\phi} \mathcal{L}_{CE} (g_\phi(f_\theta(x), y),
\end{equation*}
where $\mathcal{L}_{CE}$ indicates the cross-entropy loss. In this work we only consider the former option for its significant better performance.

\noindent\textit{ii) Independently training LLM with the concept and task labels:} The text encoder $f_\theta$, the projector $p_\psi$ and the label predictor $g_\phi$ are trained seperately with ground truth concepts labels and task labels as follows:
\begin{equation*}
    \begin{aligned}
    \theta, \psi &= \argmin_{\theta, \psi} \mathcal{L}_{CE} (p_\psi(f_\theta(x)),c), \\
    \phi &= \argmin_{\phi} \mathcal{L}_{CE} (g_{\phi}(c),y).
    \end{aligned}
\end{equation*} 
During inference, the label predictor will use the output from the projector rather than the ground-truth concepts.

\noindent\textit{iii) Sequentilally training LLM with the concept and task labels:} We first learn the concept encoder as the independent training strategy above, and then use its output to train the label predictor:
\begin{equation*}
    \begin{aligned}
    \phi = \argmin_{\phi} \mathcal{L}_{CE} (g_{\phi}(p_\psi(f_\theta(x),y).
    \end{aligned}
\end{equation*} 

\noindent\textit{iv) Jointly training LLM with the concept and task labels:} Learn the concept encoder and label predictor via a weighted sum $\mathcal{L}_{joint}$ of the two objectives described above:
\begin{equation*}
\begin{aligned}
    \theta, \psi, \phi &= \argmin_{\theta, \psi, \phi} \mathcal{L}_{joint}(x, c, y) \\ &= \argmin_{\theta, \psi, \phi} [\mathcal{L}_{CE} (g_{\phi}(p_\psi(f_\theta(x),y) \\ &+ \gamma \mathcal{L}_{CE} (p_\psi(f_\theta(x)),c)].
\end{aligned}
\end{equation*} 
 It's worth noting that the LLM-CBMs trained jointly are sensitive to the loss weight $\gamma$. We tune the value for $\gamma$ for better performance~\citep{tan2023cbm}.

 \section{B. Implementation Detail}\label{app:implement}
 In this section, we provide more details on the implementation settings of our experiments. Specifically, we implement our framework with PyTorch~\cite{paszke2017automatic} and HuggingFace~\cite{wolf2020huggingfaces} and train our framework on a single 80GB Nvidia A100 GPU. We follow a prior work~\citep{abraham2022cebab} for backbone implementation. All backbone models have a maximum token number of 512 and a batch size of 8 (for larger LLMs such as OPT-1.3B, we reduce the batch size to 1). We use the Adam optimizer to update the backbone, projector, and label predictor according to Section~\ref{sec:setup}. The values of other hyperparameters (Table~\ref{tab:app_para} in the next page) for each specific PLM type are determined through grid search. We run all the experiments on an Nvidia A100 GPU with 80GB RAM.

 \begin{table*}[htbp]
\small
\setlength\tabcolsep{5pt}
\caption{Key parameters in this paper with their annotations and evaluated values. Note that \textbf{bold} values indicate the optimal ones.}\label{tab:app_para}
\label{tab:symbols}
\centering
\begin{tabular}{@{}cccc@{}}
\toprule
\textbf{Notations}    & \textbf{Specification} & \textbf{Definitions or Descriptions}              & \textbf{Values}                  \\ \midrule
max\_len              & -                      & maximum token number of input                     & 128 / 256 / \textbf{512}                             \\
batch\_size           & -                      & batch size                                        & 8                                \\
plm\_epoch            & -                      & maximum training epochs for LLMs and the Projector     & 20                               \\
clf\_epoch            & -                      & maximum training epochs for the linear classifier & 20                               \\

\multirow{6}{*}{lr}   & DistilBERT                   & learning rate when the backbone is DistilBERT               & 1e-3 / 1e-4 / \textbf{1e-5} / 1e-6 \\
                      & BERT                   & learning rate when the backbone is BERT               & 1e-3 / 1e-4 / \textbf{1e-5} / 1e-6   \\
                      & RoBERT                   & learning rate when the backbone is RoBERT               & 1e-3 / 1e-4 / \textbf{1e-5} / 1e-6   \\
                      & OPT-125M                   & learning rate when the backbone is OPT-125M               & 1e-3 / 1e-4 / \textbf{1e-5} / 1e-6   \\
                      & OPT-350M                   & learning rate when the backbone is OPT-350               & 1e-4 / 1e-5 / \textbf{1e-6} / 1e-7   \\
                      & OPT-1.3B                & learning rate when the backbone is OPT-1.3B            & 1e-4 / 1e-5 / \textbf{1e-6} / 1e-7    \\
$\gamma$ & -                      & loss weight in the joint loss $L_{joint}$                    & 1 / 3 / \textbf{5} / 7 / 10            \\ \bottomrule
\end{tabular}
\end{table*}

\section{C. Details to Solve the Optimization Task}\label{app: optim}
As a common practice, we approximate the Hessian at $\bm{w}$ via a dampened empirical Fisher information matrix~\citep{hassibi1992second}:
\begin{equation}
    \bm{H}_\mathcal{L}(\bm\theta) \simeq \hat{\bm{F}}(\bm\theta) = \zeta\bm{I} + \frac{1}{m}\sum_{i = 1}^{m} \nabla \mathcal{L}_i (\bm\theta) \nabla \mathcal{L}_i^{\top}(\bm{\theta}),
\end{equation}
where $\zeta > 0$ is a small damplening constant, $\bm{I}$ is the indentity matrix. $m$ is the number of gradient outer products used to approximate the Hessian. 

Following~\citet{kurtic2022optimal}, based on Eq.~\eqref{eq:optimization}, we express the system of $|\bm{Q}|$ equality constrains in matrix as:
\begin{equation}
    \bm{E}_{\bm{Q}} \Delta \bm{\theta} + \bm{E}_{\bm{Q}} \bm{\theta}^\ast = 0,
\end{equation}
where $\bm{E}_{\bm{Q}} \in \mathbb{R}^{\bm{Q}\times b}$ is a matrix composed of the corresponding canonical basis vectors $\bm{e}_b (\forall b \in |\bm{Q}|)$. Using Lagrange multipliers, we hope to find stationary points of the Lagrangian $L(\Delta \bm{\theta}, \bm\lambda)$, where $\bm{\lambda} \in \mathbb{R}^{|\bm{Q}|}$ denotes a vector of Lagrange multipliers. Then, we need to solve the following system of equations:
\begin{equation}\label{eq:lagrange}
\begin{aligned}
    \frac{\partial L (\Delta \bm{\theta}, \bm{\lambda})}{\partial \Delta \bm{\theta}} = 0, \quad\quad
    \frac{\partial L (\Delta \bm{\theta}, \bm{\lambda})}{\partial \bm{\lambda}} = 0,
\end{aligned}
\end{equation}
which gives the following optimal weight update:
\begin{equation}
    \Delta \bm{\theta}^\ast = -\hat{\bm{F}}^{-1}(\bm{\theta}^\ast)\bm{E}_{\bm{Q}}^{\top}(\bm{E}_{\bm{Q}}\hat{\bm{F}}^{-1}(\bm{\theta}^\ast)\bm{E}_{\bm{Q}}^{\top})^{-1}\bm{E}_{\bm{Q}}\bm\theta^{\ast}.
\end{equation}
It prunes a set of weights $\bm{Q}$ and updates the remaining weights to preserve the loss. The corresponding loss increase incurred by the optimal weight update $\Delta\bm{\theta}^{\ast}$ can be represented as:
\begin{equation}
    \rho_{\bm{Q}} = \frac{1}{2}(\bm{E}_{\bm{Q}}\bm{\theta}^\ast)^\top (\bm{E}_{\bm{Q}}\hat{\bm{F}}^{-1}(\bm{\theta}^{\ast})\bm{E}_{\bm{Q}}^{\top})^{-1} \bm{E}_{\bm{Q}} \bm{\theta}^{\ast}.
\end{equation}
We use this as the importance score to rank groups of weights for pruning.

\section{D. Decision Pathways for Real-world Examples}
\begin{figure*}[htbp]
% \vspace{-0.2cm}
  \centering
  \scalebox{0.6}{
                  \includegraphics[width=\linewidth]{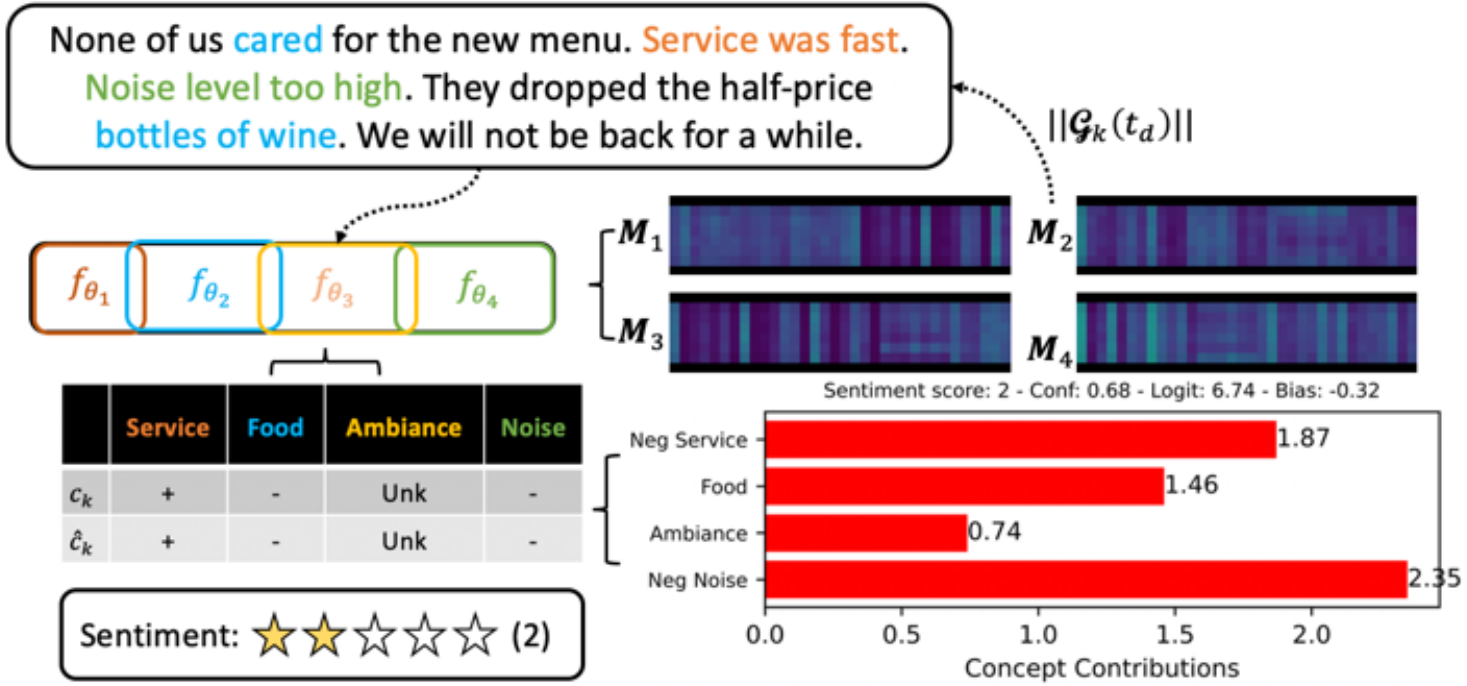}}
  % \vspace{-1mm}
  \caption{{The illustration of a decision pathway of an real-world example (\texttt{CEBaB} dataset) from the SparseCBM framework with BERT as the backbone. The binary weight masks for each concept is represented as a heatmap.}}\label{fig:example1}
  % \vspace{-4mm}
\end{figure*}

\begin{figure*}[htbp]
% \vspace{-0.2cm}
  \centering
  \scalebox{0.6}{
                  \includegraphics[width=\linewidth]{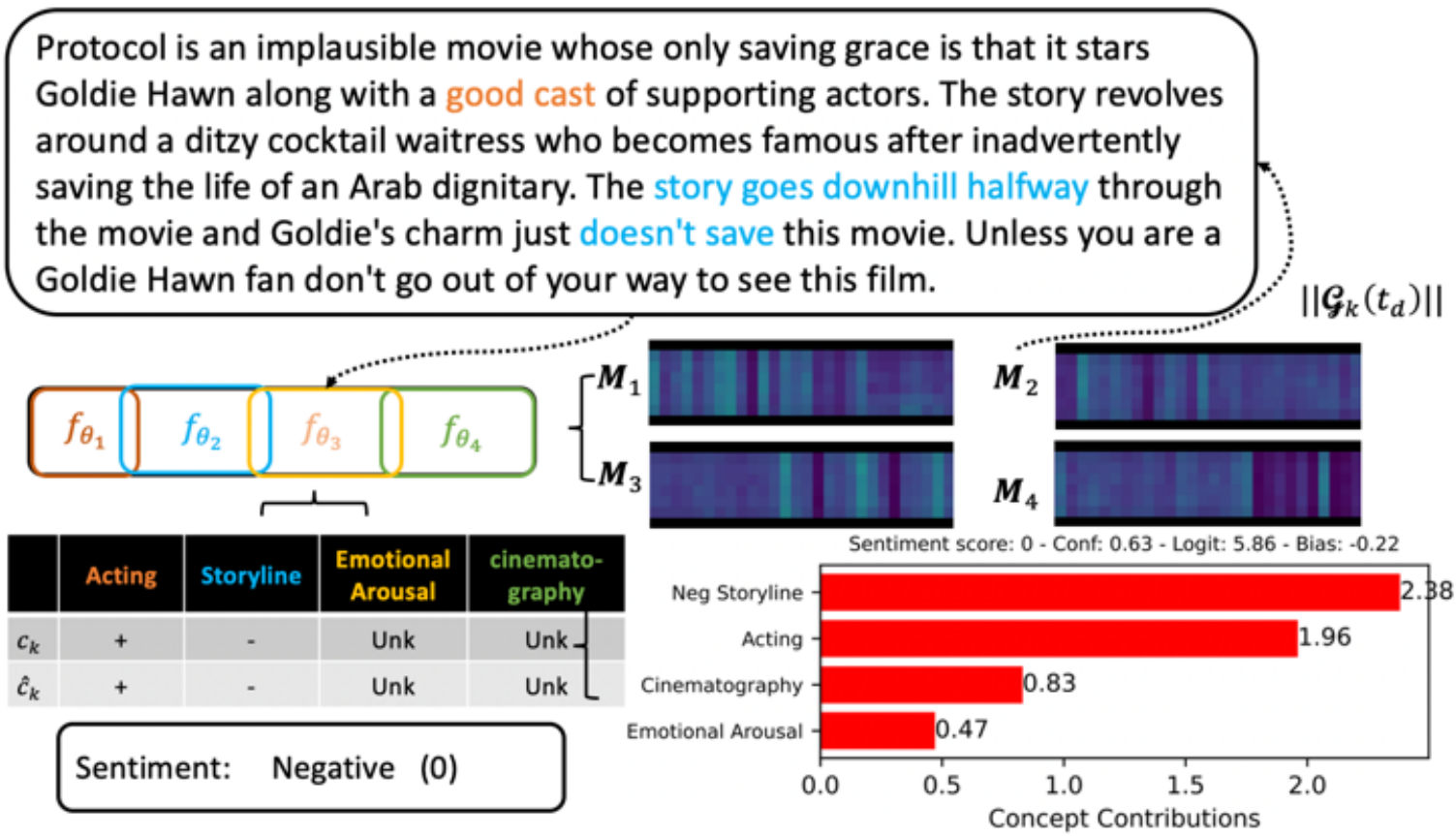}}
  % \vspace{-1mm}
  \caption{{The illustration of a decision pathway of an real-world example (\texttt{IMDB-C} dataset) from the SparseCBM framework with BERT as the backbone. The binary weight masks for each concept is represented as a heatmap.}}\label{fig:example2}
  % \vspace{-4mm}
\end{figure*}

An example from the \texttt{CEBaB} dataset is given in Figure~\ref{fig:example1} in the next page. An example from the \texttt{IMDB-C} dataset is given in Figure~\ref{fig:example2} in the next page.

\end{document}